\documentclass{article} %
\usepackage{iclr2017_conference,times}
\usepackage{hyperref}
\usepackage{url}
\usepackage{booktabs}       %
\usepackage{amsfonts}       %
\usepackage{nicefrac}       %
\usepackage{microtype}      %
\usepackage{amsmath}
\usepackage{amssymb}
\usepackage{graphicx} %
\usepackage{algorithm}
\usepackage{algorithmic}
\usepackage{array}
\usepackage{caption}
\usepackage{subcaption}
\usepackage{wasysym}
\usepackage{stackengine}


\title{A backward pass through a CNN \\using a generative model of its activations}

\author{Huayan Wang, Anna Chen, Yi Liu, Dileep George, D. Scott Phoenix\\
Vicarious\\
San Francisco, CA, USA \\
\texttt{\{huayan,anna,yiliu,dileep,scott\}@vicarious.com}
} 

\begin{document}

\maketitle

\begin{abstract}
  Neural networks have shown to be a practical way of building a very complex mapping between a pre-specified input space and output space. For example, a convolutional neural network (CNN) mapping an image into one of a thousand object labels is approaching human performance in this particular task. However the mapping (neural network) does not automatically lend itself to other forms of queries, for example, to detect/reconstruct object instances, to enforce top-down signal on ambiguous inputs, or to recover object instances from occlusion. One way to address these queries is a \emph{backward pass} through the network that fuses top-down and bottom-up information. In this paper, we show a way of building such a backward pass by defining a generative model of the neural network's activations. Approximate inference of the model would naturally take the form of a backward pass through the CNN layers, and it addresses the aforementioned queries in a unified framework.
\end{abstract}

\section{Introduction}
\label{sec:intro}
\vspace{-.1in}  
Contemporary neural networks (such as the CNNs \cite{krizhevsky2012imagenet} \cite{simonyan2013deep} \cite{szegedy2015going}) stack many layers of computations to approximate a highly complex function. In this paper we consider a \emph{backward pass} for the neural network, \emph{i.e.} some computation that initiates from a higher layer of the network and propagates information backwards. An example of the backward pass is the back-propagation algorithm for computing the gradient of the cost function. There are other forms of backward pass such as deconv-net \cite{zeiler2014visualizing}, guided-backprop \cite{springenberg2014striving}, auto-encoders \cite{HinSal06} \cite{kingma2013auto} \cite{zhao2015stacked}, hierarchical rectified Gaussians \cite{peiyun16}, and inverting representations \cite{mahendran2015understanding}.  Generally speaking, a backward pass is used to reason about some higher-layer activations of the network by mapping back to the model parameter space or input space.

Consider the following types of queries to a CNN. {\bf (1)} When given an input image that the CNN classifies as \emph{airplane}, can we ask the system to show where the airplane is (by either its segmentation mask or reconstruction)?. When there are multiple airplanes, the system should be able to realize that and find the airplanes one-by-one. {\bf (2)} When given an ambiguous input image, for example, of a piece of sheep-shaped cloud. A CNN would not classify it as sheep. But can we ask the system to ``imagine'' a sheep that fits the input? Humans are good at this. {\bf (3)} When given an input image with a heavily occluded familiar object, can we ask the system to ``imagine'' the missing part that is compatible with the visible part in various attributes (such as color, style, \emph{etc.}). {\bf (4)} When no input image is given, can we ask the system to ``imagine'' a certain class of objects (\emph{e.g.}, airplanes) and their commonly seen variations?

There is a shared theme among all the above queries: we start from a high-level concept (\emph{e.g.}, airplane or sheep), and trace it down the visual hierarchy to the image space, with different bottom-up inputs in case~1, 2 and 3, or empty bottom-up inputs in case~4. This suggests that it is highly desirable if we could tackle these questions in a unified framework.

Known ways of backward pass for CNNs are very limited in their ability to reason about object instances. For example, guided-backprop \cite{springenberg2014striving} would attribute the classification decision to contours and key features in the image without attending to a single object instance. Existing pretrained-CNN-based methods \cite{mahendran2015understanding} \cite{simonyan2013deep} for reconstructing object categories without bottom-up input is able to recover key distinctive features in a fragmented fashion. However the results are far from answering the question of what does \emph{an object instance} of that given category look like. This is expected because the CNN had no direct access to examples of single object instances in training.

One way to achieve these goals is to have a backward pass derived from a generative model for object instances. Researchers have tried to bridge CNNs and generative models in various settings \cite{NIPS2014_5423} \cite{dosovitskiy2015learning} \cite{xie2016theory}. A practical bottleneck of the generative models \cite{tu2002image} \cite{tu2005image} \cite{zia2015towards} is the inference problem, \emph{i.e.}, given an input image with clutter, how to generate an object instance to fit the one in observation without having to solve a very large search problem. Our approach is to build a hierarchical generative model for CNN activations. On each layer, we model the activations conditioned on its upper layer. Because the bottom-up mappings of the CNN are many-to-one (the \emph{invariance} property), the backward pass accommodates the variability by a one-to-many mapping. We achieve this by introducing hidden variables at each layer. In the backward pass, the hidden variables are assigned layer-by-layer in a greedy fashion, thus providing a framework to achieve efficient inference by exploiting the invariance properties of the hierarchy. Details of our approach are presented in Sec.~\ref{sec:approach}. 

We show that our model captures natural variations of objects by taking samples from it. We also show that performing a backward pass discovers object instances in a cluttered scene as modes in its posterior distribution. And we can parse a scene by repeatedly performing backward pass and subtracting the resulting object from each layer of the CNN activations.

\subsection{Related Work}

Our approach has drawn some ideas from what-where auto-encoders \cite{zhao2015stacked}. One key difference is that our approach does not use the un-pooling switches from lateral connections to account for the variability in top-down mapping. Instead, we use hidden variables on each layer allowing the variabilities to be modeled by prior distributions. This has two main benefits: (a) it allows sampling from the model without bottom-up input; (b) it allows parsing a cluttered scene---it is unclear how would the un-pooling switches from forward pass react to the presence of clutter and occlusion.

\cite{dosovitskiy2015learning} showed that a CNN can be trained to generate 3D object instances and properly relate their variations in the hidden space. However it is not clear how to perform inference given a cluttered scene. A main difference between our approach and \cite{dosovitskiy2015learning} is that we constrain the top-down activations to be similar to bottom-up activations from a pre-trained CNN, making it possible to perform efficient inference by combining bottom-up and top-down information at each layer. The hierarchical mixture model formulation also allows our model to automatically discover attributes without pre-specifying them as in \cite{dosovitskiy2015learning}.

Our approach provides an alternative view of the ``attention'' mechanism in a visual hierarchy---we attend to different objects in a scene by selecting modes in the posterior distribution. This is different from that of end-to-end trainable RNN-based approaches \cite{gregor2015draw} \cite{ba2014multiple} \cite{mnih2014recurrent} \cite{eslami2016attend}. The main difference is that our approach is more flexible in incorporating additional top-down signals, whereas the RNN-based approaches provide no easy access to its decision making process. In other words, it is not clear how to handle top-down queries such as \emph{``find the largest object''}, \emph{``find the leftmost object''}, or \emph{``find the darkest object''} without being re-trained end-to-end in each case. We will discuss this in more details in Sec.~\ref{sec:scene-parsing}.

\section{Approach}
\label{sec:approach}
\vspace{-.1in}  
\subsection{Representation}
\label{sec:rep}
\vspace{-.1in}  
Our model is a generative model of CNN activations. Consider two sets of activations: bottom-up and top-down. Let $\mathbf{F}_{bu} = \{\mathbf{F}^0_{bu}, \mathbf{F}^1_{bu}, \cdots{}, \mathbf{F}^L_{bu}\}$ be bottom-up activations on all layers of a CNN, where $\mathbf{F}^0_{bu}$ is the input image, and $\mathbf{F}^L_{bu}$ is the final classifier layer without the soft-max. Let $\mathbf{F}_{td} = \{\mathbf{F}^0_{td}, \mathbf{F}^1_{td}, \cdots{}, \mathbf{F}^L_{td}\}$ be top-down activations, where $\mathbf{F}^0_{td}$ is the reconstruction of one object instance, and $\mathbf{F}^L_{td}$ is a one-hot vector encoding the category of that object instance.

Note that all $\mathbf{F}_{bu}$ and $\mathbf{F}_{td}$ are treated as observed variables by our model: we obtain them by applying a trained CNN on a cluttered image and a masked object instance for $\mathbf{F}_{bu}$ and $\mathbf{F}_{td}$, respectively. How to design and train the CNN has profound impact in various aspects of our generative model too. We will further discuss this in Section~\ref{sec:experiment} and Section~\ref{sec:conclude}.

To account for the invariant (many-to-one) bottom-up mapping of the CNN, we would need hidden variables to help define the one-to-many top-down mappings. We denote hidden variables on layers as $\mathbf{H} = \{\mathbf{H}^0, \mathbf{H}^1, \cdots{}, \mathbf{H}^{L}\}$. They encode attributes of objects such as shape, appearance, view condition, location \emph{etc}.

We model the joint distribution by:
\begin{equation}
  P(\mathbf{F}_{bu}, \mathbf{F}_{td}, \mathbf{H}) = P(\mathbf{H}) P(\mathbf{F}_{td} \:|\: \mathbf{H}) P(\mathbf{F}_{bu} \:|\: \mathbf{F}_{td})
  \label{eq:model}
\end{equation}

\begin{figure}[h]
  \begin{center}
    \includegraphics[scale=0.2]{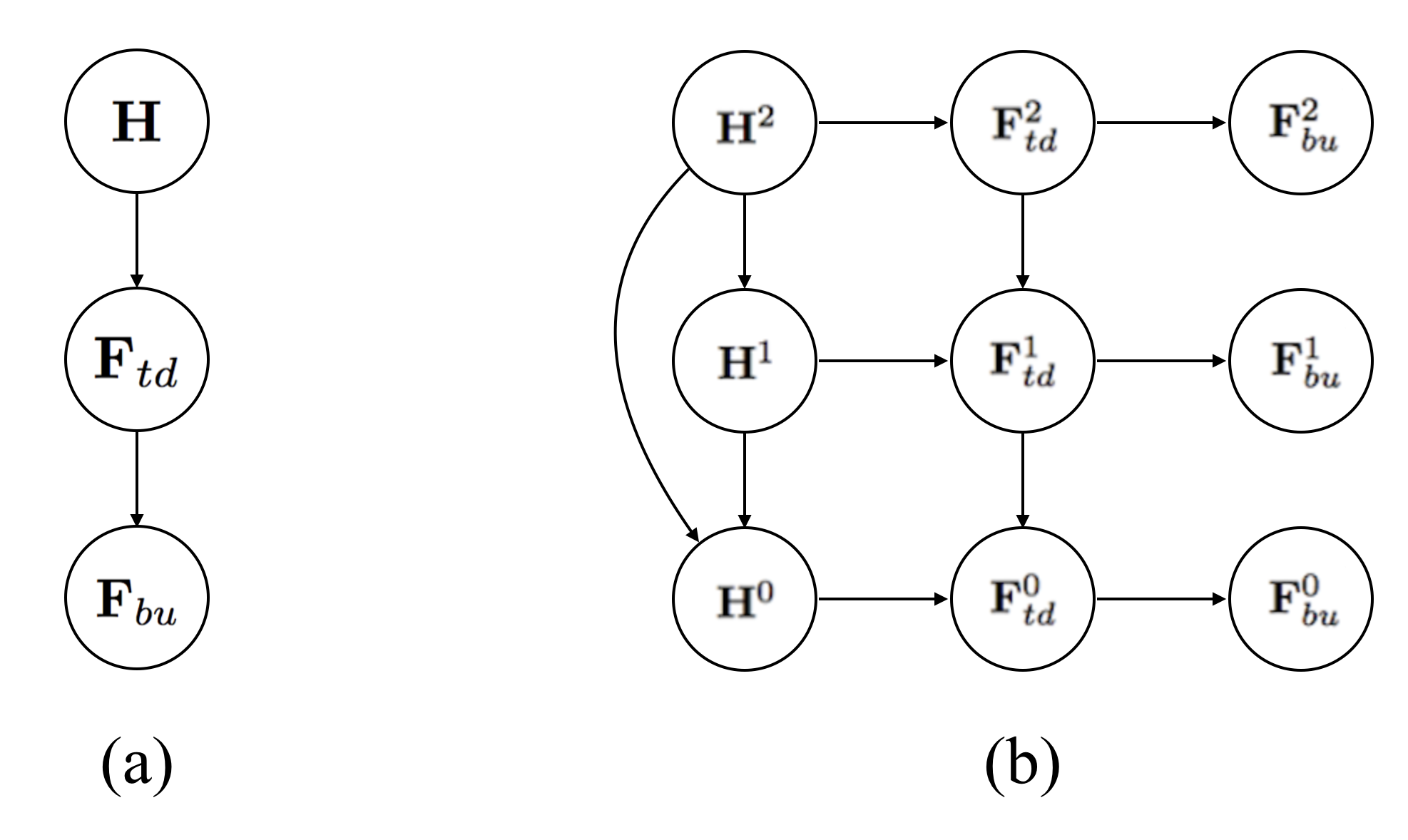}
  \end{center}
\caption{Our model as Bayesian nets. {\bf(a)} is encapsulating variables (of each type) on all layers into a single node. {\bf(b)} is separating variables on each layer into different nodes for a 3-layer hierarchy.}
\label{fig:bn}
\end{figure}

If we break it down into layers, the model can be expressed as:
\begin{equation}
  P(\mathbf{F}_{bu}^L \:|\: \mathbf{F}_{td}^L) P(\mathbf{F}_{td}^L \:|\: \mathbf{H}^L) P(\mathbf{H}^L) \prod_{l=0}^{L-1} P(\mathbf{F}_{bu}^l \:|\: \mathbf{F}_{td}^l) P(\mathbf{F}_{td}^l \:|\: \mathbf{F}_{td}^{l+1}, \mathbf{H}^l) P(\mathbf{H}^l \:|\: \mathbf{H}^{(l+1):})
  \label{eq:model-layerwise}
\end{equation}
where $\mathbf{H}^{(l+1):} = \{\mathbf{H}^{l+1}, \mathbf{H}^{l+2}, \cdots{}, \mathbf{H}^{L}\}$. Figure~\ref{fig:bn} shows our model represented as Bayes nets.

When the model is trained, we can reason about object instances and their attributes by the conditional distribution $P(\mathbf{F}_{td}, \mathbf{H} \:|\: \mathbf{F}_{bu})$: different object instances would correspond to different modes in the posterior. And we can select (``attend to'') them in a top-down manner by clamping (or imposing a prior on) the hidden attributes $\mathbf{H}$. We will elaborate on this in Sec.~\ref{sec:scene-parsing}. We can also sample from $P(\mathbf{F}_{td}, \mathbf{H})$ to visualize the variations of object instances captured by the model.

\subsubsection{Top-down generation (attribute instantiation)}
\label{sec:prior}

\vspace{-.1in}
In this subsection we introduce the top-down generation part of our model:
\begin{equation}
P(\mathbf{F}_{td}, \mathbf{H}) = P(\mathbf{F}_{td}^L \:|\: \mathbf{H}^L) P(\mathbf{H}^L) \prod_{l=0}^{L-1} P(\mathbf{F}_{td}^l \:|\: \mathbf{F}_{td}^{l+1}, \mathbf{H}^l) P(\mathbf{H}^l \:|\: \mathbf{H}^{(l+1):})
  \label{eq:prior}
\end{equation}

The top layer activation $\mathbf{F}^L_{td}$ is a one-hot vector with each dimension corresponding to an object category. We use $P(\mathbf{F}^L_{td} \:|\: \mathbf{H}^L) P(\mathbf{H}^L)$ to encode a prior distribution over categories. In this work we simply assume $\mathbf{F}^L_{td} = \mathbf{H}^L$, so by clamping $\mathbf{H}^L$ (or specifying $P(\mathbf{H}^L)$) we can issue queries such as ``attend to an airplane''. In principle we could allow encoding arbitrary distributions over categories conditioned on $\mathbf{H}^L$ and handle queries such as ``attend to something you would find in a kitchen''. For more sophisticated applications, it could also provide an interface to some higher level reasoning mechanism which issues queries to the visual hierarchy.

The remaining terms describe how activations on a layer are generated by its upper layer activations and hidden variables. The general idea is to use a mixture model of top-down convolutions, \emph{i.e.} using $K$ sets of top-down convolution weights and choose among them by a hidden index $\gamma \in \{0,\cdots{},K-1\}$.

The hidden variable has two parts: $\mathbf{H}^l = \{ \gamma^l, \mathbf{\delta}^l\}$, where $\gamma^l$ is the mixture component index, and $\mathbf{\delta}^l = \{\delta^l_y, \delta^l_x\}$ is a spatial offset applied to the output nodes. In this work we use the same $\gamma$ and $\mathbf{\delta}$ for all nodes on the same layer $l$. More expressiveness could be attained by relaxing this constraint and possibly adding spatial lateral connections among them.

Given the hidden attributes $\mathbf{H}^l$, we then instantiate them into feature activations on the lower layer by:
\begin{equation} 
  P(\mathbf{F}^l_{td} \:|\: \mathbf{F}^{l+1}_{td}, \mathbf{H}^l) = \prod_i P(f^{l,i}_{td} \:|\: \mathbf{F}^{l+1}_{td}, \mathbf{H}^l)
\end{equation}
\begin{equation}
P(f^{l,i}_{td} \:|\: \mathbf{F}^{l+1}_{td}, \mathbf{H}^l) \sim \mathcal{N}(\phi(\mathbf{F}^{l+1}_{td}, \mathbf{H}^l)(i), \sigma^l_0)
\end{equation}
where $f^{l,i}_{\star}$ denotes the $i$'th node of $\mathbf{F}^l_{\star}$; $\phi(\mathbf{F}^{l+1}_{td}, \mathbf{H}^l)$ is the top-down mixture-of-convolutions function that applies top-down filters to $\mathbf{F}^{l+1}_{td}$ according to the mixture component index and the offsets, parameterized by $K$ sets of convolution filters:
\begin{equation}
  \phi(\mathbf{F}^{l+1}, \gamma, \delta_y, \delta_x; \: \{\mathbf{W}^l_k, \mathbf{b}^l_k\}_{k=1,\cdots{},K}) = \textnormal{ReLU}(\textnormal{conv2d}(\mathbf{F}^{l+1}, \mathbf{W}^l_{\gamma}) + \mathbf{b}_{\gamma}) \circ \delta_y \circ \delta_x
  \label{eq:top_down_conv_offset}
\end{equation}
We use $\circ \: \delta_{y/x}$ denotes applying a vertical/horizontal offset to the feature map.

The last term $P(\mathbf{H}^l \:|\: \mathbf{H}^{(l+1):})$ captures ``consistency'' of hidden variable assignments. In our current model
\begin{equation}
P(\mathbf{H}^l \:|\: \mathbf{H}^{(l+1):}) = P(\mathbf{\delta}^l)P(\gamma^l \:|\: \gamma^{(l+1):})
\end{equation}
where $P(\mathbf{\delta}^l)$ is a uniform distribution over the domain of $\mathbf{\delta}^l$ (that we describe how to choose in experiment section). The other term $P(\gamma^l \:|\: \gamma^{(l+1):})$ captures consistency of mixture-component selections in consecutive layers. We can think of it as a layer-specific ``n-gram model'' \cite{brown1992class}. For example, if the order of the n-gram model is $n$, the choice of mixture component at layer $l$ is conditioned on a ``prefix'' of length $n$ consisting of the mixture component indices on the previous (upper) $n$ layers. Specifically we use a conditional multinomial distribution with Dirichlet prior.

\subsubsection{Observation model}
\label{sec:likelihood}
\vspace{-.1in}

The observation model $P(\mathbf{F}_{bu} \:|\: \mathbf{F}_{td})$ is specified by element-wise conditional Gaussian:
\begin{equation}
  P(\mathbf{F}_{bu} \:|\: \mathbf{F}_{td}) = \prod_{l,i} P(f^{l,i}_{bu} \:|\: f^{l,i}_{td})
\end{equation}
The object-instance-activation $f^{l,i}_{td}$ and the scene-activation $f^{l,i}_{bu}$  at the corresponding node are related by:
\begin{equation}
P(f^{l,i}_{bu} \:|\: f^{l,i}_{td}) \sim \begin{cases}
  \mathcal{N}(f^{l,i}_{td}, \sigma^l_1), & \text{if } f^{l,i}_{td}\geq \lambda^l \\
  \mathcal{N}(\beta^l, \sigma^l_2), & \text{otherwise}
\end{cases}
\label{eq:elem_likelihood}
\end{equation}
where $\lambda^l$ is a layer-specific activation threshold: for each node, if the single object would activate it beyond $\lambda^l$, the scene's activation is a Gaussian around the object's activation; otherwise the scene activation is determined by a background Gaussian centered at a constant $\beta^l$.

The parameters $(\lambda^l, \beta^l, \sigma^l_1, \sigma^l_2)$ reflect the characteristics of the CNN. For example, $\sigma^l_1$ can be viewed as a measure of the robustness of the representation with respect to background clutter. One would naturally expect it to vary across feature channels. Our inference and learning methods work no matter if $\sigma^l_1$ is feature-channel-specific or feature-channel-independent. In experiments we found that using feature-channel-specific $\sigma^l_1$'s makes the backward pass more robust to clutter.

\subsection{Inference}
\label{sec:infer}
\vspace{-.1in}  
\subsubsection{Approximate MAP Inference}
\label{sec:infer-map}
\vspace{-.1in}

A query to reason about object instances in a scene takes the form:
\begin{equation}
 \underset{\mathbf{F}_{td}, \mathbf{H}}{\text{maximize}}\: P(\mathbf{F}_{td}, \mathbf{H} \:|\: \mathbf{F}_{bu})
  \label{eq:map}
\end{equation}

We could also clamp some variable in $\mathbf{H}$ (usually in high layers) for selecting object instances in the scene in a top-down manner. This will be further discussed in Section~\ref{sec:experiment}.

Solving (\ref{eq:map}) exactly is intractable because the underlying graphical model has high tree-width. However, if we assign the variables in a greedy manner\footnote{The quality of this approximation is related to compositionality of the CNN's feature representation. We further discuss this in supplementary material.} from upper layers to lower layers, each step is readily solvable. This top-down procedure will take the form of a \emph{backward pass} through the CNN layers.

The approximation amounts to solving the following subproblem for each layer in top-down order:
\begin{equation}
  \underset{\mathbf{F}_{td}^l, \mathbf{H}^l}{\text{maximize}} \: P(\mathbf{F}_{bu}^l | \mathbf{F}_{td}^l) P(\mathbf{F}_{td}^l | \mathbf{F}_{td}^{l+1}, \mathbf{H}^{l}) P(\mathbf{H}^{l} | \mathbf{H}^{(l+1):})
  \label{eq:layer-map}
\end{equation}
Note that, given the choice of $\mathbf{H}^l$, the maximization can be done independently for each element of $\mathbf{F}_{td}^l$. That is, the above maximization is equivalent to
\begin{equation}
  \underset{\mathbf{H}^l}{\text{maximize}} \: P(\mathbf{H}^{l} | \mathbf{H}^{(l+1):}) \prod_i \max_{f^{l,i}_{td}} P(f^{l,i}_{bu} | f^{l,i}_{td}) P(f^{l,i}_{td} | \mathbf{F}^{l+1}_{td}, \mathbf{H}^{l})
  \label{eq:max_outer_loop}
\end{equation}
Solving this amounts to iterating over all choices of $\mathbf{H}^l$, and for each fixed $\mathbf{H}^l$ solve the following problem for each $f^l_i$ independently:
\begin{equation}
\underset{f^{l,i}_{td}}{\text{maximize}} \: P(f^{l,i}_{bu} | f^{l,i}_{td}) P(f^{l,i}_{td} | \tilde{f}^{l,i})
\label{eq:max_single_elem}
\end{equation}
where $\tilde{f}^{l,i}$ is the $i$-th element of the top-down convolutional feature map $\phi(\cdot{})$ that we defined in (\ref{eq:top_down_conv_offset}).

According to the model specification we have
\[
P(f^{l,i}_{bu} | f^{l,i}_{td}) P(f^{l,i}_{td} | \tilde{f}^{l,i}) = \begin{cases}
  \mathcal{N}(f^{l,i}_{bu}; f^{l,i}_{td}, \sigma^l_1) \cdot{} \mathcal{N}(f^{l,i}_{td}; \tilde{f}^{l,i}, \sigma^l_0), \quad & \text{if } f^{l,i}_{td} \geq \lambda^l  \\
  \mathcal{N}(f^{l,i}_{bu}; \beta, \sigma^l_2) \cdot{} \mathcal{N}(f^{l,i}_{td}; \tilde{f}^{l,i}, \sigma^l_0), \quad & \text{otherwise}
\end{cases}
\]
where $\mathcal{N}(x;\mu,\sigma)$ denotes the Gaussian density function with parameters $(\mu, \sigma)$ evaluated $x$. We can simply solve the two cases and take the maximum. On first range $f^{l,i}_{td} \geq \lambda$, consider the weighted mid-point $\eta = ((\sigma^l_0)^2 f^{l,i}_{bu} + (\sigma^l_1)^2 \tilde{f}^{l,i}) / ((\sigma^l_0)^2 + (\sigma^l_1)^2)$: we have the maximizer at $\eta$ if $\eta \geq \lambda^l$, and $\lambda^l$ otherwise. On the second range $f^{l,i}_{td} < \lambda^l$, we have the maximizer at $\tilde{f}^{l,i}$ if $\tilde{f}^{l,i} < \lambda^l$, and $\lambda^l$ otherwise\footnote{The reader may have noticed that the boundary value $\lambda^l$ can only be attained for the first range but not the second. However, in practice we allow $f^{l,i}_{td}$ to attain $\lambda^l$ on both ranges in maximization. This is mathematically rigorous if we slightly change the definition of $P(f^{l,i}_{bu} | f^{l,i}_{td})$ at the boundary. For brevity we put this in supplementary material.}. Finally we take the maximum over the two solutions.

\subsubsection{Sampling}
\vspace{-.1in}  
When no bottom up input is given, we can sample from the prior (\ref{eq:prior}) by the following steps. Sample from $P(\mathbf{F}_{td}^L \:|\: \mathbf{H}^L) P(\mathbf{H}^L)$ to choose a category-of-interest. For each layer from $L-1$ down to $0$: sample hidden variables from $P(\mathbf{H}^l \:|\: \mathbf{H}^{(l+1):})$ and then sample from $P(\mathbf{F}^l_{td} | \mathbf{F}^{l+1}_{td}, \mathbf{H}^{l})$. To sample from $P(\mathbf{F}^l_{td} | \mathbf{F}^{l+1}_{td}, \mathbf{H}^{l})$ we compute the top-down convolution with given filter index and hidden offset; and add a Gaussian noise.

\subsection{Learning}
\label{sec:learning}
\vspace{-.1in}

Because $\mathbf{F}_{td}$ is fully observed in our model, the learning problem decomposes into learning $P(\mathbf{F}_{bu} \:|\: \mathbf{F}_{td})$ and learning $P(\mathbf{H}) P(\mathbf{F}_{td} \:|\: \mathbf{H})$ respectively.

\subsubsection{Observation model}
The observation model $P(\mathbf{F}_{bu} \:|\: \mathbf{F}_{td})$ is parameterized by $\lambda^l, \beta^l, \sigma_1^l, \sigma_2^l$ describing, for each element in the feature map, how does its activation from a scene relates to that from the single object been modeled. To fit these parameters we collect a dataset of cluttered scenes with masks for each object. Then we compute $\mathbf{F}_{bu}$ by applying the CNN to the scene image and compute $\mathbf{F}_{td}$ by applying the CNN to the masked object instances. The Gaussian parameters $\beta^l, \sigma_1^l, \sigma_2^l$ can be easily computed for any given threshold value $\lambda^l$, and we maximize the likelihood function by enumerating different values of $\lambda^l$ chosen densely from an interval.

\subsubsection{Hard EM}

To learn the parameters in $P(\mathbf{H}) P(\mathbf{F}_{td} \:|\: \mathbf{H})$ we use the hard EM algorithm. Again we use images with object instance masks to compute $\mathbf{F}_{td}$. In the E-step we impute all hidden variables in $\mathbf{H}$. That amounts to solving the layer-wise inference problem (\ref{eq:layer-map}) without the term involving $\mathbf{F}^l_{bu}$. In the M-step we try to maximize all model parameters given the fixed hidden variables. Parameters in $P(\mathbf{H})$ can be fitted in closed form. (They are just CPD tables.) Parameters in the top-down convolutions need to be maximized by back-propagation and SGD. Instead of running SGD to convergence in each M-step, we only run it for a (small) fixed number of epochs.

\section{Experiments}
\label{sec:experiment}
\vspace{-.1in}  
\begin{figure}[t!]
  \vspace*{-.1in}  
\begin{center}
  \includegraphics[width=0.1\textwidth]{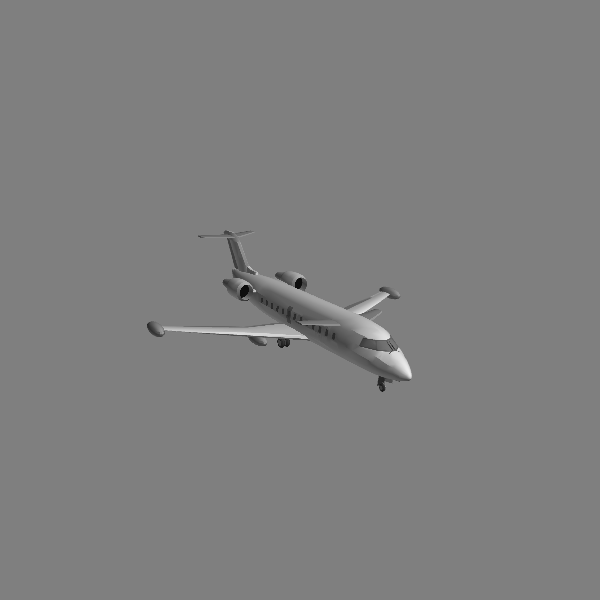}
  \includegraphics[width=0.1\textwidth]{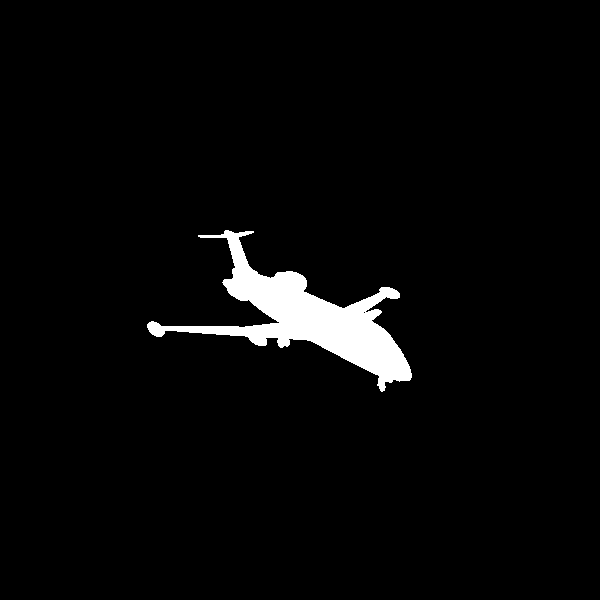}
  \hspace{0.2cm}
  \includegraphics[width=0.1\textwidth]{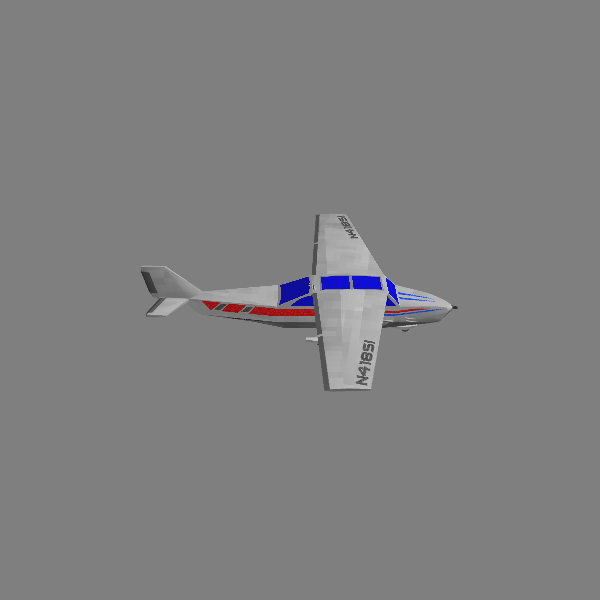}
  \includegraphics[width=0.1\textwidth]{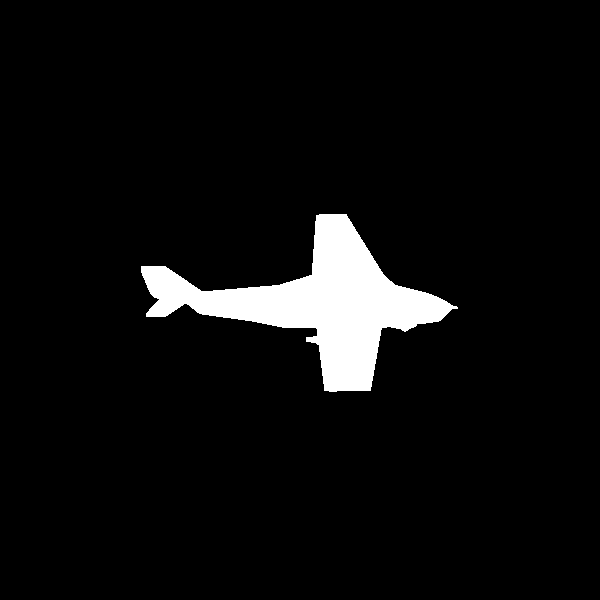}
  \hspace{0.2cm}
  \includegraphics[width=0.1\textwidth]{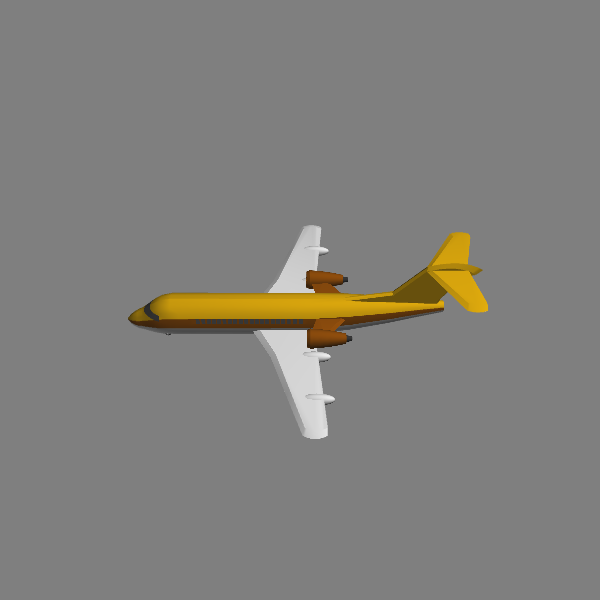}
  \includegraphics[width=0.1\textwidth]{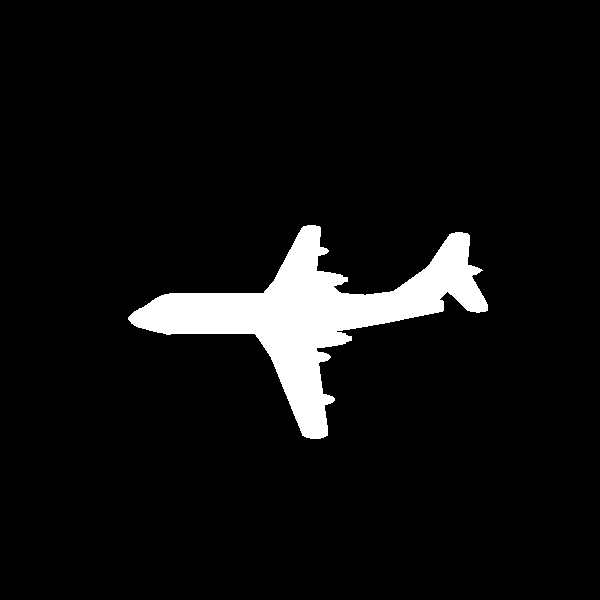}
  \hspace{0.2cm}
  \includegraphics[width=0.1\textwidth]{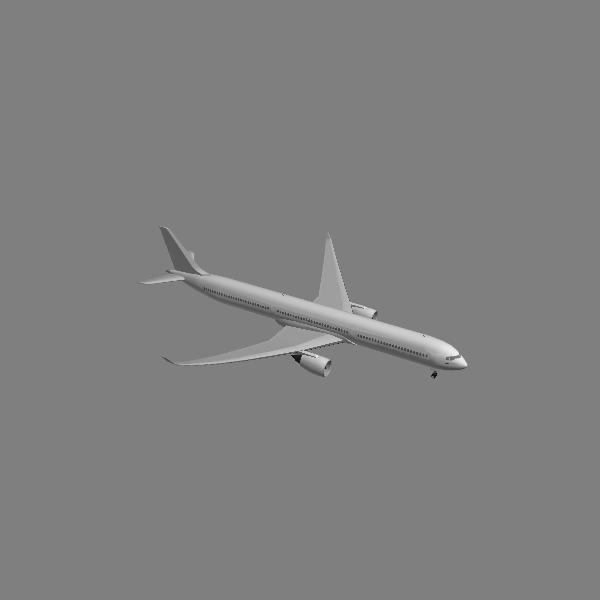}
  \includegraphics[width=0.1\textwidth]{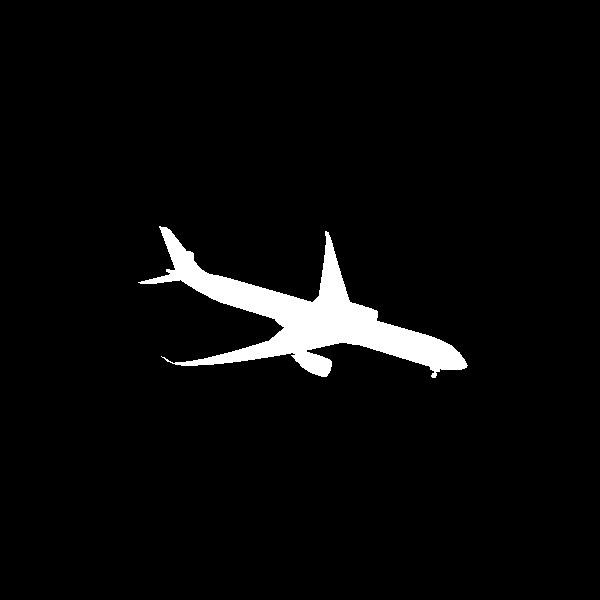}
  \end{center}
\vspace*{-.1in}  
  \caption{Example images from our rendered dataset of airplanes on clean background with mask.}
  \label{fig:dataset}
\end{figure}

We build our model on the pre-trained VGG network \cite{simonyan2014very}. In the architecture of VGG, the feature map after the fifth pooling layer (\emph{pool5}) is a tensor with shape $512\times7\times7$ (512 channels of $7\times 7$ feature map). In forward pass, this tensor is flattened into a vector and fed into two fully-connected layers and the soft-max classifier. The fully-connected layers are tailored for the classification task but not amenable to our model. We explain this in detail by demonstrating feature compositionality of VGG in supplementary material. To that end we discard the fully-connected layers and treat \emph{pool5} as $\mathbf{F}^{L-1}$. We choose the range of hidden offsets $(\delta_y, \delta_x)$ to be $[-2,2]\times[-2,2]$ for \emph{pool5}; $[-1,1]\times[-1,1]$ for all other pooling layers; and $\{0\}\times\{0\}$ (fixed) for all convolutional layers. We collected a dataset of 3,600 images of airplanes on clean background with mask obtained from rendering CAD models (Fig.~\ref{fig:dataset}). We used 3,000 of these images for training and the rest for creating test scenes. Although we have only trained for one category, this is already sufficient for demonstrating the key characteristics of our model (Sec.~\ref{sec:scene-parsing}). We will scale up our model to more categories and larger training sets in future work.

\subsection{Sampling From Prior}
\vspace{-.1in}  
\begin{figure}[H]
  \begin{center}
        \vspace*{-.1in}  
  \includegraphics[width=0.1\textwidth]{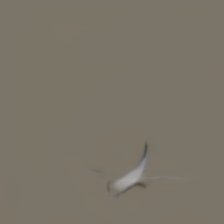}%
  \hspace{0.05cm}
  \includegraphics[width=0.1\textwidth]{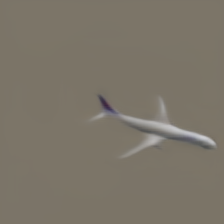}%
  \hspace{0.05cm}
  \includegraphics[width=0.1\textwidth]{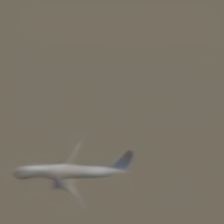}%
  \hspace{0.05cm}
  \includegraphics[width=0.1\textwidth]{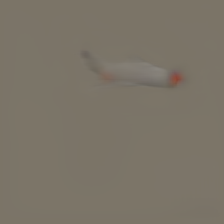}%
  \hspace{0.05cm}
  \includegraphics[width=0.1\textwidth]{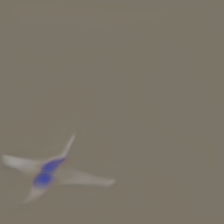}%
  \hspace{0.05cm}
  \includegraphics[width=0.1\textwidth]{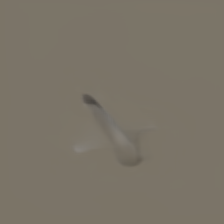}%
  \hspace{0.05cm}
  \includegraphics[width=0.1\textwidth]{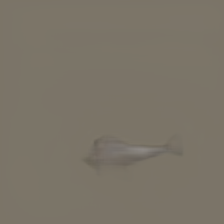}%
  \hspace{0.05cm}
  \includegraphics[width=0.1\textwidth]{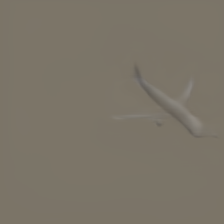}%
  \hspace{0.05cm}
  \includegraphics[width=0.1\textwidth]{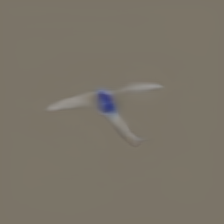}%
\end{center}
\vspace*{-.1in}  
\caption{Samples from our model. Better viewed when enlarged on a screen.}
\label{fig:free-samples}
\end{figure}
\begin{figure}[H]
    \includegraphics[width=0.1\textwidth]{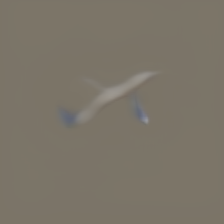}%
    \includegraphics[width=0.1\textwidth]{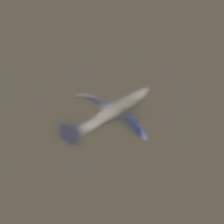}%
    \includegraphics[width=0.1\textwidth]{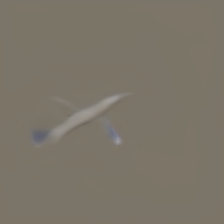}%
    \includegraphics[width=0.1\textwidth]{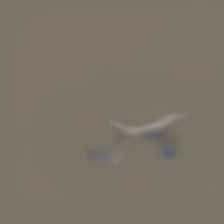}%
    \includegraphics[width=0.1\textwidth]{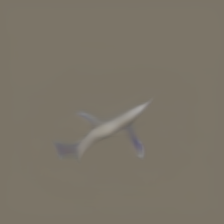}
    \hspace{0.01cm}
    \includegraphics[width=0.1\textwidth]{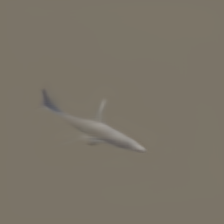}%
    \includegraphics[width=0.1\textwidth]{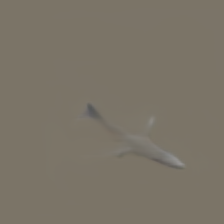}%
    \includegraphics[width=0.1\textwidth]{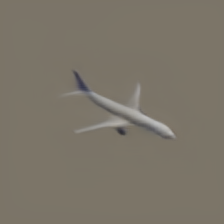}%
    \includegraphics[width=0.1\textwidth]{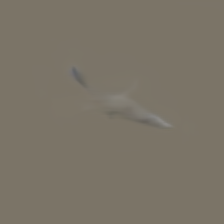}%
    \includegraphics[width=0.1\textwidth]{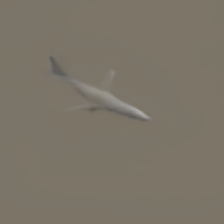}\\
    \includegraphics[width=0.1\textwidth]{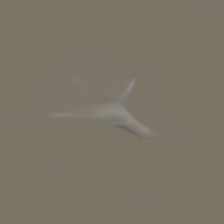}%
    \includegraphics[width=0.1\textwidth]{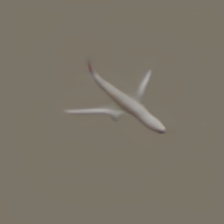}%
    \includegraphics[width=0.1\textwidth]{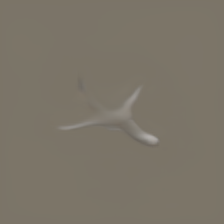}%
    \includegraphics[width=0.1\textwidth]{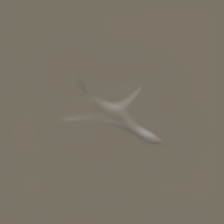}%
    \includegraphics[width=0.1\textwidth]{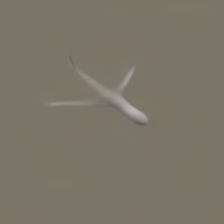}
    \hspace{0.01cm}
    \includegraphics[width=0.1\textwidth]{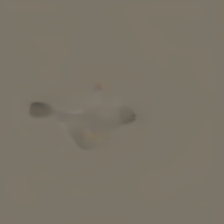}%
    \includegraphics[width=0.1\textwidth]{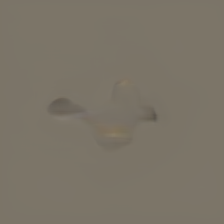}%
    \includegraphics[width=0.1\textwidth]{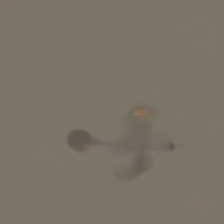}%
    \includegraphics[width=0.1\textwidth]{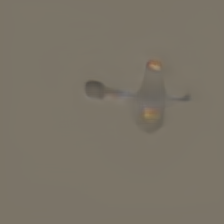}%
    \includegraphics[width=0.1\textwidth]{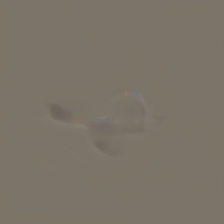}\\
    \includegraphics[width=0.1\textwidth]{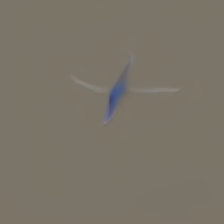}%
    \includegraphics[width=0.1\textwidth]{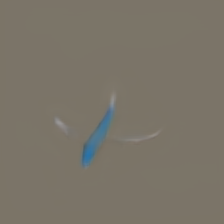}%
    \includegraphics[width=0.1\textwidth]{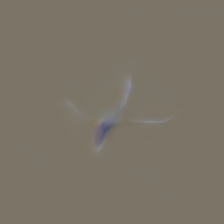}%
    \includegraphics[width=0.1\textwidth]{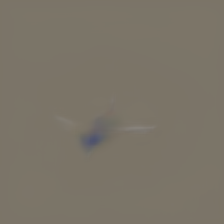}%
    \includegraphics[width=0.1\textwidth]{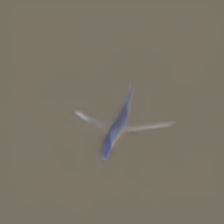}
    \hspace{0.01cm}
    \includegraphics[width=0.1\textwidth]{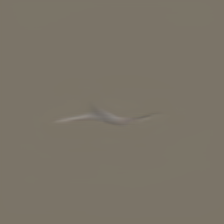}%
    \includegraphics[width=0.1\textwidth]{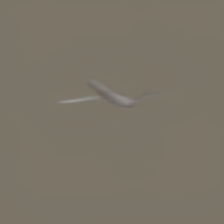}%
    \includegraphics[width=0.1\textwidth]{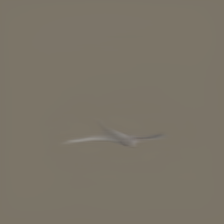}%
    \includegraphics[width=0.1\textwidth]{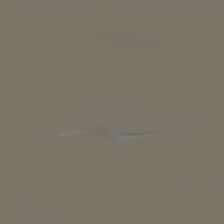}%
    \includegraphics[width=0.1\textwidth]{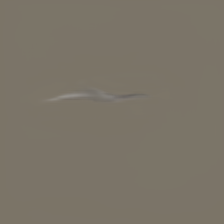}
\vspace*{-.2in}      
    \caption{Samples by clinching to a hidden assignments on the second-to-top layer. See text for more explanations. Better viewed when enlarged on a screen.}
    \label{fig:clinched-samples}
\end{figure}
In Fig.~\ref{fig:free-samples} we show some samples from our model\footnote{Technically speaking a sample from our model contains activations on all layers of the hierarchy. Only the activations in image space are shown.}. We can see that it has captured many variations of planes. In Fig.~\ref{fig:clinched-samples} we show samples when we clinch the hidden variables at the second-to-top layer, \emph{i.e.} \emph{pool5} of VGG. (The top layer has the one-hot vector encoding the category ``airplane''.) That is, each group of five samples share the same $(\gamma, \delta_y, \delta_x)$ on the second-to-top layer. Comparing these to Fig.~\ref{fig:free-samples} we can see that most of the variations are encoded at the second-to-top layer. We had to choose $K^l=500$ for the second-to-top layer and $K^l=5$ for all other layers to attain reasonable training error on each layer. This reflects a characteristics of the underlying CNN. Contemporary CNNs are trained as a classifier without much constraint on its internal representations. Naturally they try to keep most of the information all the way up to the latest layers \cite{mahendran2015understanding} \cite{dosovitskiy2015inverting} and transform it to be amenable to the classification task. However an ideal representation for our model to build on should be one where invariance is accumulated gradually through each layer of the hierarchy. In that case not only does our model capture variations in a more elegant way, its efficiency in MAP inference would also be boosted. We will further discuss this in Sec.~\ref{sec:conclude}.

\subsection{MAP Inference and Scene Parsing}
\label{sec:scene-parsing}
\begin{figure}[H]
    \vspace*{-.1in}
  \begin{center}
\includegraphics[width=.9\textwidth]{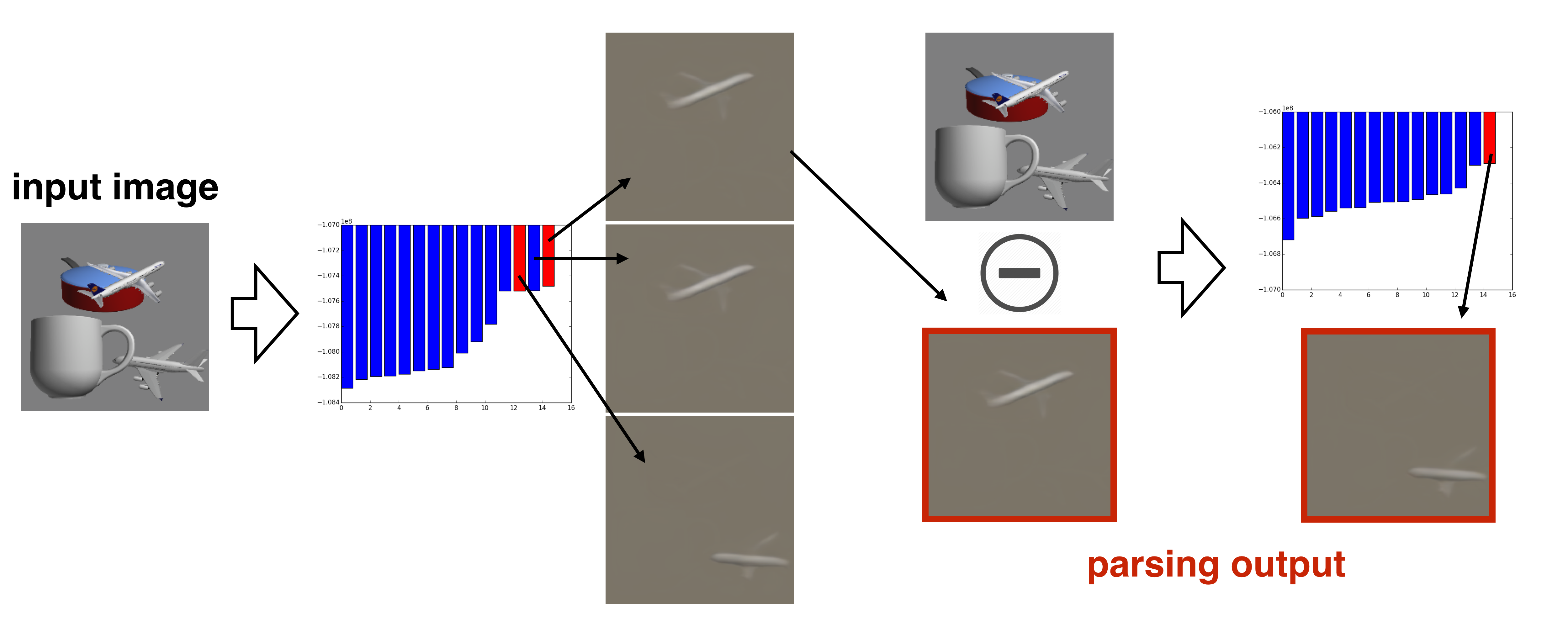}
  \end{center}
  \vspace*{-.1in}      
  \caption{Using backward pass repeatedly to parse all instances of a given category. At each step, different object instances correspond to different modes in the posterior distribution of our model. See text for more explanations. Better viewed when enlarged on a screen. The red bars are ``true hypotheses''---posterior modes corresponding to the two airplanes in the scene.}
  \label{fig:parsing}
\end{figure}

\vspace{-.1in}  
In this section we demonstrate MAP inference of our model and how it can be embedded into an iterative scene parsing algorithm. On the left of Fig.~\ref{fig:parsing} we have a rendered test image with two planes and some clutter. The two planes are from the held out test set. We will refer them as top-plane and bottom-plane. The first backward pass would find the MAP solution corresponding to the top-plane. To visualize the posterior distribution we backtraced the top $15$ assignments on layer \emph{pool5} and showed their log probabilities in a bar plot. We can see that the bottom-plane also corresponds to a prominent mode in the posterior. Then we take the MAP solution and subtract it from the bottom-up feature maps on all layers of the hierarchy (not only subtracting in the image space). The next backward pass would have a new MAP solution corresponding to the bottom-plane as shown on the right of Fig.~\ref{fig:parsing}. This procedure can be repeated for any number of steps to parse a scene. And we can also improve its robustness by maintaining multiple parsing paths (sequences of object hypotheses).
As we have mentioned in Sec.~\ref{sec:intro}, this parsing procedure has very different characteristics than the RNN-based approaches such as A.I.R.~\cite{eslami2016attend}. The latter was featured by end-to-end training, which brought about some benefits, but also led to the blackbox nature of its decision making in each parsing step. For example, as demonstrated in \cite{eslami2016attend}, the RNN sometimes learns a spatial policy that scans from one side of the image to the other; sometimes learns a category based policy that always discovers different categories of objects in a fixed order. The network ``automatically chooses'' these policies without an interface for top-down intervention. When we have no control over the scanning policy, the expected time complexity of discovering any given object in a scene is linear in the total number of objects therein. This is markedly different from how humans do it---our scene parsing can easily incorporate top-down queries such as \emph{``find the largest object''}; \emph{``find the leftmost object''}, or \emph{``find the darkest object''} without task-specific training.

To that end, our approach is more amenable to top-down signals in the form of constraints / prior distributions over hidden variables or the top-layer activation of the model. For example, in the first parsing step of the test case above, we could specify a preference over location (\emph{i.e.} $(\delta_y, \delta_x)$ on \emph{pool5}) and easily choose between the two planes. If we had trained the model for multiple categories, we could choose which category to attend to by specifying $\mathbf{F}^{L}$ at the top. If we have learned the model with the hidden indices $\gamma^l$ associated to interpretable attributes, we could also specify our preference over those attributes in the backward pass. This property could be quite useful if the visual hierarchy were to serve as a part of a system that does higher level reasoning with the learned representations. To fully demonstrate this aspect of our approach will be our future work.

\subsection{Real Images}
\label{sec:realimage}
\vspace{-.1in}  
\begin{figure}
  \vspace*{-.1in}
\begin{center}
  \includegraphics[width=0.2\textwidth]{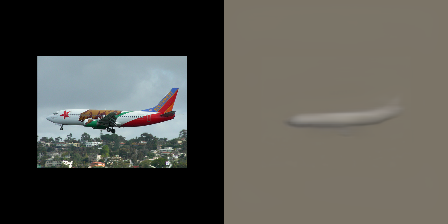}
  \includegraphics[width=0.2\textwidth]{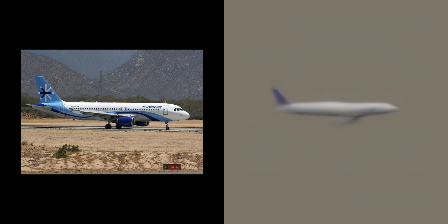}
  \includegraphics[width=0.2\textwidth]{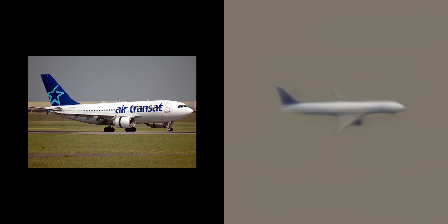}
  \includegraphics[width=0.2\textwidth]{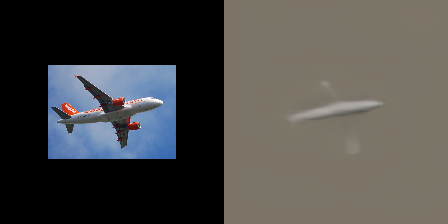}\\
  \includegraphics[width=0.2\textwidth]{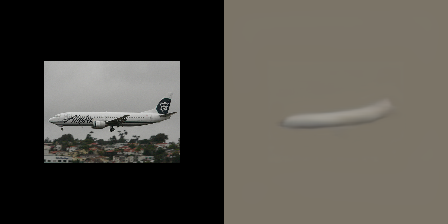}
  \includegraphics[width=0.2\textwidth]{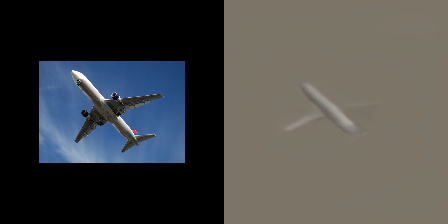}
  \includegraphics[width=0.2\textwidth]{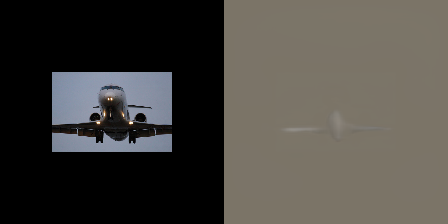}
  \includegraphics[width=0.2\textwidth]{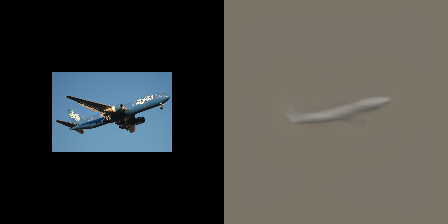}
\end{center}
\vspace*{-.1in}
  \caption{Example results on real images. For each pair, left is the input image and right is the MAP inference result in image space. Better viewed when enlarged on a screen.}
  \label{fig:realimage}
\end{figure}
Our model, albeit trained on rendered planes, works well on real images as shown in Fig.~\ref{fig:realimage}. One issue to notice is that variations in scale has not been capture in our model, so we had to scan over scales in inference (by scaling and padding the input image). A principled solution would be to introduce dedicated hidden variables for scale variations. That requires the underlying CNN to encode scale invariance in proper manner. We will address this in future work.

\section{Discussions and Future Work}
\label{sec:conclude}
\vspace{-.1in}
Our model currently uses a single shared $\gamma^{l}$ for all nodes on the same layer. In reality, attributes are better characterized locally rather than globally, and they are usually conditioned on parent node attributes. This observation has two implications. One is that we should allow using different $\gamma^{l}$ for different elements of the feature map on the same layer, \emph{i.e.} apply different filters at different locations in the top-down convolution. In that case the mixture component indices become a tensor $\mathbf{\gamma}^{l} = \{\gamma^{l,i}\}$, and we would also introduce a prior term $P(\mathbf{\gamma}^l)$ that can be viewed as spatial laterals enforcing the consistency between choices at neighboring spatial locations.

The other implication is that we cannot have the hierarchy keep all information at the top since there is no global factorization of \emph{all} attributes. Notably, as we have observed from Fig.~\ref{fig:free-samples} and Fig.~\ref{fig:clinched-samples}, and also demonstrated by other researchers \cite{mahendran2015understanding} \cite{dosovitskiy2015inverting}, contemporary CNNs retain most of the visual attributes at its top layers (and refactor them in a way that is amenable to the classification task). However, an ideal hierarchical visual representation from our perspective is one where invariances to various visual attributes are gradually acquired layer by layer. The generative model for backward pass would then encode these attributes as hidden variables in corresponding layers, so that they are assigned in the backward pass by incorporating top-down and bottom-up information. A better ``spread'' of invariance throughout the hierarchy would also benefit the time complexity of backward pass (MAP inference). Let $K^l$ be the size of joint hidden state space at layer $l$. Note that the total amount of variation captured by the hierarchy is $O(\prod_l K^l)$, whereas the time complexity of backward pass is $O(\sum_l K^l)$. Obviously, efficiency of the hierarchy is maximized by having similar $K^l$ on all layers. That corresponds to having invariance gradually accumulated in forward pass. How to train CNNs with these characteristics is also a topic of our future work.

{\small
\bibliography{references}
\bibliographystyle{iclr2017_conference}
}

\section*{Supplementary Material}

\subsection*{Feature compositionality}

Feature compositionality is a property of a feature representation that \emph{the representation of the whole is composed of the representations of the parts.} This property is usually not attained in the internal representation of systems that had been trained end-to-end for a specific task. However, this is a highly desirable property if we were to make use of the learned representation in a different task from the one in training (just like in this paper). Feature compositionality is also manifested in the knowledge representation of humans as we interpret visual scenes as objects and parts.

\begin{figure}[hb]
  \begin{center}
    \includegraphics[width=0.8\textwidth]{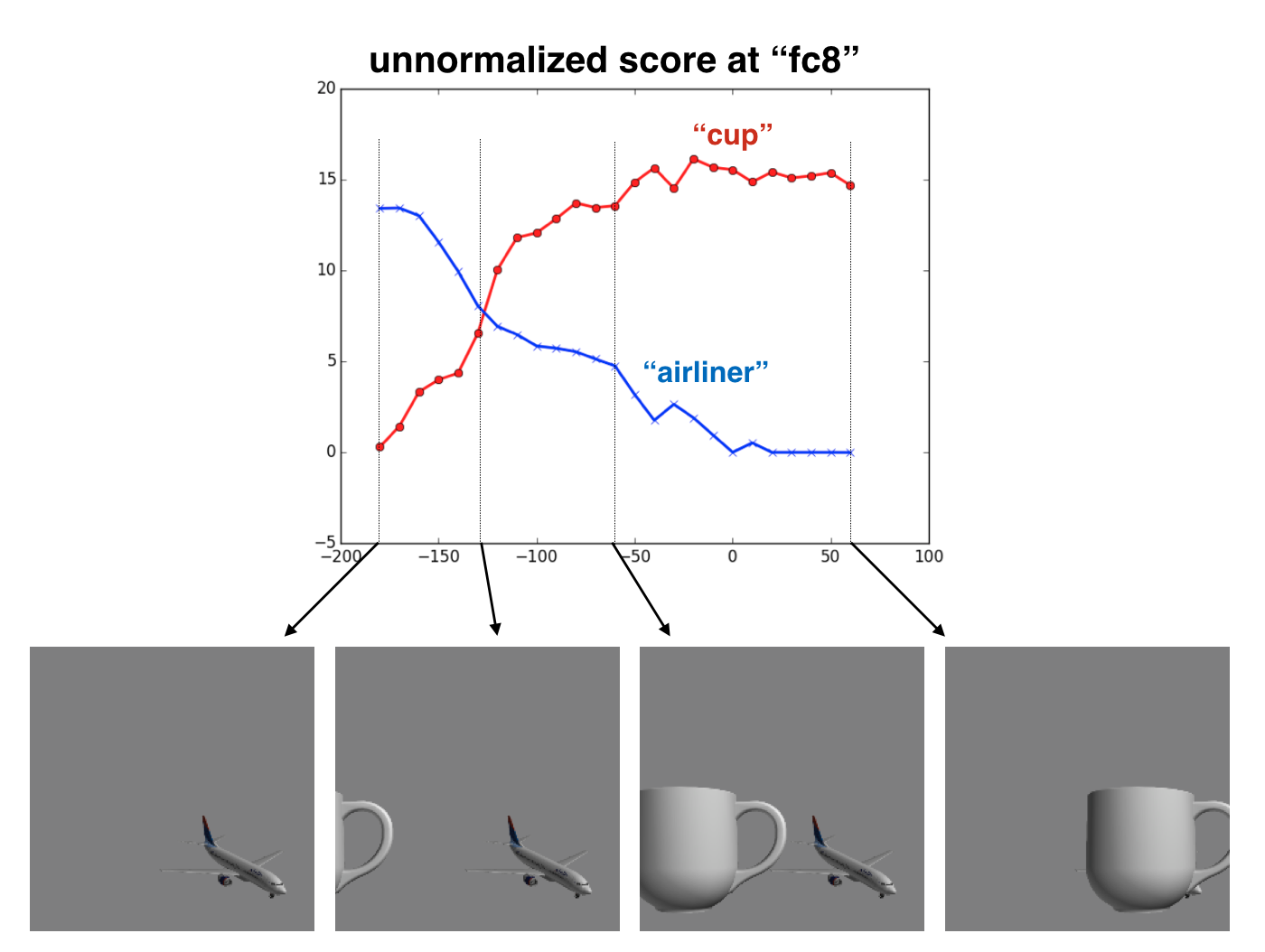}
\end{center}
  \caption{Examining feature compositionality on layer \emph{fc8} of VGG. See text for explanations.}
  \label{fig:feature-compositionality-fc8}
\end{figure}

To investigate feature compositionality of VGG, we created a synthetic sequence (Fig.~\ref{fig:feature-compositionality-fc8}) where a plane is in a fixed position and a distractor object is moved in from the left ($10$ pixels per frame) until it completely occludes the plane. On each frame we apply VGG forward pass and examine the classifier score at \emph{fc8}. Note that this is the layer before soft-max, so scores do not necessarily suppress each other. In Fig.~\ref{fig:feature-compositionality-fc8} we are plotting the score for \emph{cup} versus \emph{airliner} (both are among original ImageNet labels). Four key frames corresponding to the four vertical lines in the plot are shown under the figure. The key observation here is that the score for ``airliner'' has dropped dramatically long before the distractor gets anywhere close to the plane.

\begin{figure}
  \begin{center}
    \includegraphics[width=0.4\textwidth]{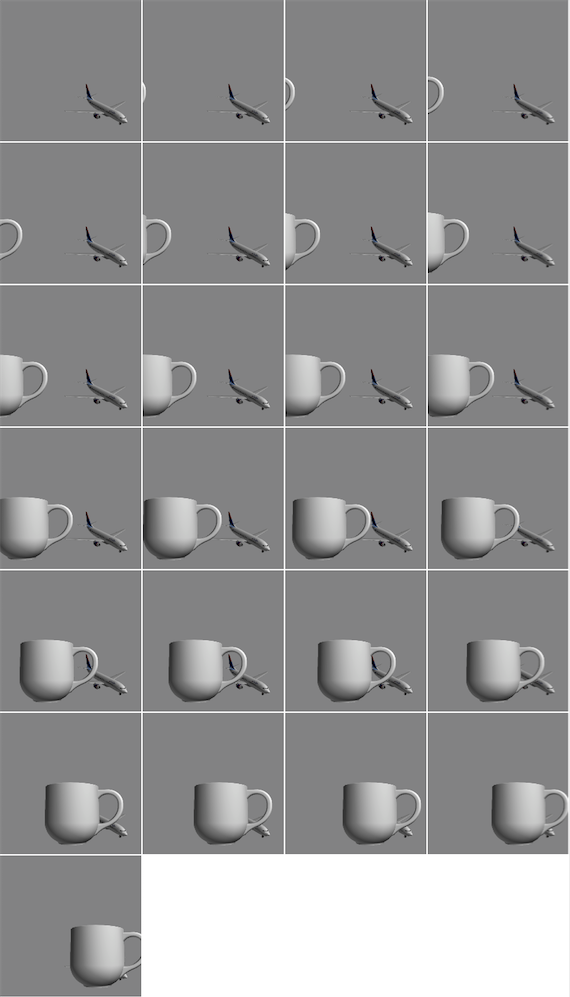}
    \includegraphics[width=0.4\textwidth]{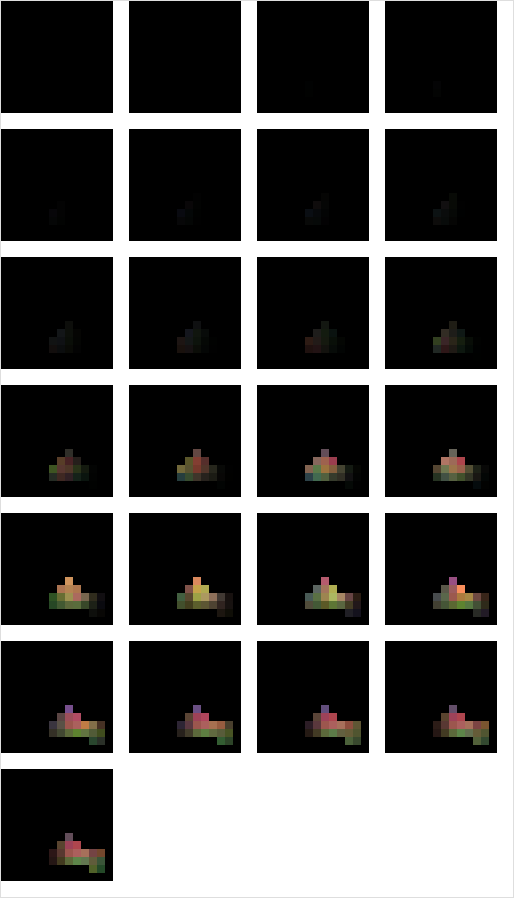}
\end{center}
  \caption{Examining feature compositionality on the layer \emph{conv5-3} of VGG. See text for explanations.}
  \label{fig:feature-compositionality-conv}
\end{figure}

In fact, it is natural for the fully connected layers to have this effect given the original single label training objective of the CNN. However, one might wonder if this is entirely due to the fully connected layers. To examine that we also visualized in Fig.~\ref{fig:feature-compositionality-conv} the feature compositionality on \emph{conv5-3}, which the topmost convolutional layer of VGG with a shape of $512\times 14\times 14$.

On the left hand side of Fig.~\ref{fig:feature-compositionality-conv} we are showing each frame of the synthetic sequence. On the right hand side of Fig.~\ref{fig:feature-compositionality-conv} we are visualizing, for each frame, the change of activations on the layer \emph{conv5-3} from that from the clean plane image. The change is a $512\times 14\times 14$ tensor. We map it into RGB by randomly splitting the 512 channels into 3 groups and maximizing over each group. To help visualization we also masked it using the mask the of plane projected to $14\times 14$. The intensity of the pixels reflect the magnitude of change, and they share the same scale (therefore comparable) across different frames. We can see that even when the distractor and the plane are separate, the change of activations of the plane is reaching a magnitude similar to the case where the distractor completely occludes the plane. 

In our backward pass, to make reliable decisions at higher layers when the object been modeled is embedded in a cluttered scene, it is highly desirable that the CNN forward pass satisfies feature compositionality to a certain degree, such that we can relate the feature activations from the single object ($\mathbf{F}_{td}$) and that from the scene ($\mathbf{F}_{bu}$). This ``robustness to clutter'' is captured by parameter $\sigma_1$ in our model. It turns out that in VGG some feature channels are more robust to clutter than others. By using feature-channel-specific $\sigma_1$ we managed to get reasonable results with clutter as shown in Sec.~\ref{sec:scene-parsing}.

\subsection*{Rigorous definition of likelihood term}
In order for the closed-form maximization method of (\ref{eq:max_single_elem}) to work rigorously, we need to re-define the boundary behavior of
the likelihood term (\ref{eq:elem_likelihood}):
\[
P(f_{bu} | f_{td}, s) \sim \begin{cases}
  \mathcal{N}(f_{td}, \sigma_1), & \text{if } f_{td}> \lambda, \: \text{ or } f_{td} = \lambda, \; s=0. \\
  \mathcal{N}(\beta, \sigma_2), & \text{if } f_{td} < \lambda, \: \text{ or } f_{td} = \lambda, \; s=1 \\
  \end{cases}
\]
That is, at the boundary $f_{td}= \lambda$ we toss a coin $s$ to decide which side to go. And $s$ is introduced as a hidden variable in model by giving it a uniform prior distribution over $\{0,1\}$.
In approximate MAP inference, we always maximize over $s$, that gives rise the closed-form solution that we described in Sec.~\ref{sec:infer-map}.

\end{document}